\documentclass[letterpaper]{article}
\usepackage{aaai}
\usepackage{times}
\usepackage{helvet}
\usepackage{courier}
\usepackage[show]{chato-notes} 

\usepackage{graphicx}
\usepackage{url}
\usepackage{mathptmx}
\usepackage{hyphenat}
\usepackage{amssymb}
\usepackage{amsmath}

\usepackage{balance}
\usepackage{subfig} 

\usepackage{booktabs} 

\newcommand{\sectionrule}{\addlinespace[0.5ex]}

\usepackage{pifont}
\newcommand{\cmark}{\ding{52}} 
\newcommand{\xmark}{\ding{56}} 

\newcommand{\hide}[1]{}

\usepackage{cmm-greek}

\makeatletter
\renewcommand{\paragraph}{%
  \@startsection{paragraph}{4}%
  {\z@}{1.00ex \@plus 1ex \@minus .2ex}{-1em}
  {\normalfont\normalsize\bfseries}%
}
\makeatother

\begin{document}

\title{
How to Ask for a Favor: A Case Study on the Success of Altruistic Requests
}

\author{
Tim Althoff$^*$, Cristian Danescu-Niculescu-Mizil$^\dagger$, Dan Jurafsky$^*$\\
$^*$Stanford University, $^\dagger$Max Planck Institute SWS\\
\tt \footnotesize
althoff$\vert$jurafsky@stanford.edu, cristian@mpi-sws.org
}

\maketitle

\begin{abstract}
Requests are at the core of many social media systems such as question \& answer sites and online philanthropy communities. 
While the success of such requests is critical to the success of the community, the factors that lead community members to satisfy a request are largely unknown. 
Success of a request depends on factors like \emph{who} is asking, \emph{how} they are asking, \emph{when} are they asking, and most critically \emph{what} is being requested, ranging from small favors to substantial monetary donations. 
We present a case study of {\em altruistic} requests in
an online community where all requests ask for the very same contribution and do not offer anything tangible in return, allowing us to disentangle what is requested from textual and social factors.  
Drawing from social psychology literature, we extract high-level social features from text that operationalize 
social relations between recipient and donor and demonstrate that these extracted relations are predictive of success.
More specifically, we find that clearly communicating need through the narrative is essential and that linguistic indications of gratitude, evidentiality, and generalized reciprocity, as well as high status of the asker further increase the likelihood of success.
Building on this understanding, we develop a model that can predict the success of unseen requests, significantly improving over several baselines. 
We link these findings to research in psychology on helping behavior, providing a basis for further analysis of success in social media systems.

\end{abstract}

\section{Introduction}
\label{sec:intro}

We live in a time where people increasingly turn to the web for help. 
Our needs, however, often go far beyond mere information from existing webpages and we need help from real people.
For example, we ask for answers to specific questions on \emph{StackOverflow.com}, 
for 
donations on
 \emph{DonorsChoose.org},
or for help on online social communities such as \emph{Reddit.com}.  
In each of these cases a user performs a request, which we define as an act of asking formally for something.
All these communities rely heavily  on their members to help satisfy the request.
Yet, the factors that lead community members to satisfy a request remain largely unknown.  
Understanding the dynamics and factors of successful requests has the potential to substantially improve such communities by educating users about better formulating requests and promoting likely-to-succeed requests \cite{greenberg2013crowdfunding,mitra2014language}. 
In addition to these practical benefits,
understanding the factors that make a request successful 
has implications for questions in social psychology and linguistic pragmatics.

Studies on the popular crowdfunding platform Kickstarter have shown that
the success of a request depends most crucially on \emph{what} is being requested, that is, whether it is a small favor like an answer to a simple question or a large financial contribution \cite{mitra2014language,mollick2014dynamics}.
Many other factors need to be controlled as well; what the giver receives \emph{in return}, 
\emph{when} they are asking, and even group dynamics, since people are more likely
to give to projects that others are already giving to \cite{etter2013launch,ceyhan2011dynamics,mitra2014language}.
Satisfying a request on peer-to-peer lending or crowd-funding platforms
can also bring a reward, and this also can drive the selection process.
It is extremely difficult to disentangle the effects of all these factors 
in determining what makes people satisfy requests, and what makes them select some requests over others.

In this paper, we develop a framework for controlling for each of these 
potential confounds  while
studying the role of two aspects that characterize compelling requests: 
\emph{social} factors ({\em who} is asking and how the recipient is related to the donor and community)
and \emph{linguistic} factors ({\em how} they are asking and what linguistic 
devices accompany successful requests). 
With the notable exception of 
\citeauthor{mitra2014language} (\citeyear{mitra2014language}),
the effect of language on the success of requests has largely been ignored thus far.\footnote{Linguistic factors have also been considered to influence 
the response quantity, quality, and speed to questions 
in online communities and social networks \cite[inter alia]{teevan2011factors,burke2007introductions}.}

Our goal is to understand what motivates people to give when they do not receive anything tangible in return.
That is, we focus on the important special case of {\em altruistic} requests in which the giver receives no rewards.
This controls for the incentive to obtain attractive rewards commonly offered on crowdfunding sites such as Kickstarter;
the absence of external factors such as tangible rewards also makes the language itself
all the more important in persuading others to help.
In this domain we also do not need to consider crowdfunding-related marketing strategies such as emphasizing limited time offers (scarcity)
or showing that other people made the same decision already (social proof) \cite{cialdini2001influence},
which are known to manifest themselves in language \cite{mitra2014language}.
Second, we focus on requests that a single user can fulfill,
thereby additionally eliminating
group behavior effects such as herding \cite{ceyhan2011dynamics} or completing donation biases \cite{wash2013value}. 
Finally, we focus on one community in which {\em what} is being asked for is held constant.
This allows us to explore a large number of different requests of different individual users, at different times, that all have the same goal.
Controlling for the request goal therefore allows us to study how to optimize
a particular request solely by optimizing its presentation, and helps provide a direct practical
benefit to the requester (by contrast, advising a requester who needs something to instead ask for something different may be 
advice of limited practical use).

We therefore chose to study donations in ``Random Acts of Pizza'', an online community devoted to giving away free pizza to strangers that ask for one.
Random Acts of Pizza\footnote{\scriptsize\url{http://www.reddit.com/r/Random_Acts_Of_Pizza}} (RAOP) is a community within the social news and entertainment website \emph{Reddit.com}.
Users can submit requests for free pizza and if their story is compelling enough a fellow user might decide to send them one, ``\emph{because... who doesn't like helping out a stranger? The purpose is to have fun, eat pizza and help each other out. Together, we aim to restore faith in humanity, one slice at a time.}\footnote{\scriptsize\url{http://www.randomactsofpizza.com}}''
A typical post might sound something like this: 
``\emph{It's been a long time since my mother and I have had proper food. I've been struggling to find any kind of work so I can supplement my mom's social security... A real pizza would certainly lift our spirits} \cite{ABCNewsRAOP}.''

This platform addresses many of the potential confounds that complicate other platforms or studies:
all requests ask for the same thing, a pizza, there are no additional incentives or rewards,
each request is satisfied by a single user, 
users and requests are embedded in a social network within Reddit, 
and requests are largely textual.  This dataset thus
provides us with an unusually clear picture of the effect of language and social factors on success.

The remainder of this paper is organized as follows:
inspired by studies in crowdfunding, user-to-user evaluations in social networks, and helping behavior in social psychology, we introduce a variety of textual and social factors that are potentially associated with successful requests.
We use topic modeling and automatic detection to extract a particularly complex factor, the {\em narrative}  structure of requests.
We employ a logistic regression framework
to test what factors matter in the community,
showing
that narratives are significantly correlated with success,
and that
signaling gratitude, the intention to reciprocate in the future, supporting the narrative with additional evidence, 
as well as a high status of the user within the community further increase the chance of success.
We do not find any support for theories predicting that positive sentiment, politeness, and
user similarity are associated with success.
Thus, drawing from social psychology literature, our extracted high-level social features operationalize the relation between recipient and donor.
We then demonstrate in a prediction task that the proposed model generalizes to unseen requests and significantly improves over several baselines.

\section{The Dataset}
\label{sec:dataset}

Our dataset\footnote{Available at \scriptsize \url{cs.stanford.edu/~althoff/raop-dataset/}} contains the entire history of the Random Acts of Pizza Subreddit from December 8, 2010 to September 29, 2013 (21,577 posts total).
To compute user features we further crawled the entire lifetime history of posts and comments across all Subreddits for all users involved in the RAOP Subreddit (1.87M submissions total).
The community only publishes which \emph{users} have given or received pizzas but not which \emph{requests} were successful. 
In the case of successful users posting multiple times it is unclear which of the requests was actually successful.
Therefore, we restrict our analysis to users with a single request for which we can be certain whether or not it was successful, leaving us with 5728 pizza requests.
We split this dataset into development (70\%) and test set (30\%) such that both sets mirror the average success rate in our dataset of 24.6\%. 
All features are developed on the development test only while the test set is used only once to evaluate the prediction accuracy of our proposed model on held-out data.
For a small number of requests (379) we further observe the identity of the benefactor through a ``thank you'' post by the beneficiary after the successful request.
This enables us to reason about the impact of user similarity on giving.\footnote{Reddit's front page showing the popular articles can skew exposure, but this will not effect RAOP as posts generally receive about two orders of magnitude less up-votes than would be necessary to appear on Reddit's front page.}

\section{Success Factors of Requests}
\label{sec:factors}

Previous work on crowdfunding, helping behavior and user-to-user evaluations in social networks 
have pointed to a number of textual and social factors that could influence the success of a request.

\subsection{Textual Factors of Success}

\paragraph{Politeness} A person experiencing gratitude is more likely to behave prosocially towards their benefactor and others \cite{tsang2006gratitude,bartlett2006gratitude,mccullough2001gratitude}.
However, gratitude is only one component of politeness \cite{Danescu-Niculescu-Mizil+al:13b}. Other indicators include deference, greetings, indirect language, apologizing and hedges. 
We ask a more general question: does a polite request make you more likely to be successful?

\paragraph{Evidentiality} Some requests emphasize the {\em evidence} for the narrative or need.
The literature on helping behavior literature suggests that urgent requests are met more frequently than non-urgent requests \cite{yinon1987reciprocity,shotland1983emergency,colaizzi1984similarity,gore1997effects}.

\paragraph{Reciprocity} In social psychology, \emph{reciprocity} refers to responding to a positive action with another positive action.
People are more likely to help if they have received help themselves \cite{wilke1970obligation}. 
Since in altruistic domains, there is no possibility of direct reciprocity, we hypothesize that
recipients might pay the kindness \emph{forward} to another community member, a concept known as ``generalized reciprocity'' \cite{willer2013payitforward,gray2012payingitforward,plickert2007networkreciprocity}.
Feelings of gratitude can elicit this behavior \cite{gray2012payingitforward}.
We hypothesize that the community would be more willing to fulfill the request of someone who is likely to contribute to the community later on.

\paragraph{Sentiment} While many requests are fairly negative, talking about lost jobs, financial problems, or relationship breakups, some of them are positive, asking for pizza for birthday parties and other celebrations. 
Helping behavior literature predicts that positive mood is associated with a higher likelihood of giving \cite{forgas1998askingnicely,milberg1988moods}. 
While these studies refer to the sentiment or emotional state of the benefactor, the most closely related linguistic feature that is available in this setting would be the sentiment of the text. 
Thus, the literature would predict that very positive requests are more likely to succeed. 
We additionally expect that very negative requests could be more successful, too, since they most likely describe very unfortunate situations of the requester.

\paragraph{Length}
Studies on the success of research grant proposals have shown that the simple factor of request length can be significantly related to funding success even when controlling for a variety of other factors \cite{lettice2012lengtheffect}.
We hypothesize that longer requests will be interpreted as showing
more effort on the side of the requester and giving them the
opportunity to provide more evidence for their situation.

\subsection{Social Factors}
Studies on crowdfunding have shown that the size of the social network of the project creator is associated with success \cite{mollick2014dynamics,mitra2014language}.
Work on user-to-user evaluations in online social networks suggests that the success of a request depends on who you are as a user and particularly that notions of user status and user similarity could be influential in the process \cite{anderson2012effects,leskovec2010signed,guha2004propagation}.
We study both status and similarity in this work. 

\paragraph{Status}
Studies in social psychology have found that people of high status, e.g. defined by occupation or wealth, receive help more often \cite{solomon1977status,goodman1993influence}.

\paragraph{Similarity}
People are more likely to help those who resemble them \cite{colaizzi1984similarity,chierco1982effects,emswiller1971similarity}.
We predict that users will be more likely to give pizza to users who are like them in some way.

\subsection{What Narratives Drive Success?} 
The textual part of a request, the narrative, has been shown to significantly influence the outcome in peer-to-peer lending platforms
\cite{herzenstein2011tell,greenberg2013crowdfunding,mitra2014language}.
In order to understand the nature and power of different narratives
without coding them manually \cite{herzenstein2011tell},
we explore automatic methods of narrative extraction.
Consider the following two pizza requests:

\medskip

\noindent \textbf{Example 1:} \vspace{-1mm}
{\small
\begin{quote}
``My gf and I have hit some hard times with her losing her job and then unemployment as well for being physically unable to perform her job due to various hand injuries as a server in a restuarant. She is currently petitioning to have unemployment reinstated due to medical reasons for being unable to perform her job, but until then things are really tight and ANYTHING would help us out right now. 

I've been both a giver and receiver in RAOP before and would certainly return the favor again when I am able to reciprocate. It took everything we have to pay rent today and some food would go a long ways towards making our next couple of days go by much better with some food.''
\end{quote}
}

\noindent \textbf{Example 2:} \vspace{-1mm}
{\small
\begin{quote}
``My friend is coming in town for the weekend and my friends and i are so excited because we haven't seen him since junior high. we are going to a high school football game then to the dollar theater after and it would be so nice if someone fed us before we embarked :)''
\end{quote}
}

While the first request (successful) goes into detail about hard times (and claims to reciprocate) 
the second one (unsuccessful) merely aims at ``being fed''.

To identify the different kinds of stories we draw on previous literature suggesting
that narratives can be automatically extracted using topic modeling and related techniques
\cite{chambers-jurafsky:2009:ACLIJCNLP,wallace:2012:NAACL-HLT}. We therefore
perform topic modeling
through non-negative matrix factorization (NMF) \cite{hoyer2004nmf}
of a TF-IDF weighted bag-of-words representation \cite{salton1988tfidf}
of the requests in our dataset. 
We additionally enforce sparsity
on the topic distribution for each request to shape the topics in
a way that captures most of a given request, and restrict ourselves
to nouns (using the Stanford Part-Of-Speech Tagger\footnote{\scriptsize\url{http://nlp.stanford.edu/software/}}).
We choose to use 10 topics and use a SVD-based initialization
for NMF \cite{boutsidis2008svd}.

The resulting topics are shown in Table \ref{table:topic_modeling} along with descriptive names, the 15 highest-scoring terms and the success rate (fraction of requests that successfully obtained pizza).
We observe that many topic clusters follow a specific theme and that their success rates vary dramatically (the average success rate is 24.6\%). 
Topics \textsc{Money1} and \textsc{Money2} focus on money, and
the high success rate of topic \textsc{Money1} (32.3\%) suggests that this is a particularly successful narrative.
Topic \textsc{Job} is similarly successful (31.9\%) and features job related terms.
A large number of requests further seem to come from college students talking about studying for classes and finals, their roommates, and the university (topic \textsc{Student}).
Another narrative in the data are requests for and about family (topics \textsc{Time\&Family} and \textsc{Money2}). 
This narrative can be identified by the usage of words indicating family relationships like kid, husband, family, mother, wife, and parents (not all of them are included in the top 15 terms).
Topic \textsc{Friend} stands out since it noticeably worse than any other topic (17.0\%).
It captures requests asking for pizza for a friend that is in town, to cater for parties, or to provide culinary support for a night of movie watching with the girlfriend. 
We hypothesize that stories of this topic display little actual need for pizza (particularly compared to stories talking about money and job problems) and simply communicate a pizza craving by the requester.
We further recognize that many requests employ previously defined factors such as gratitude (``thanks in advance'' in topic \textsc{Gratitude}), providing pictures as additional evidence, and the intention to ``pay it forward''.

\begin{table}[t!]
\centering
\resizebox{1.0\columnwidth}{!}{
\begin{tabular}{l p{0.10\columnwidth} p{0.72\columnwidth} }
\toprule
\textbf{Name} & \textbf{SR} & \textbf{Terms}\\
\midrule
\textsc{Money1} & 32.3\% & week ramen paycheck work couple rice check pizza grocery rent anyone favor someone bill money \\
\sectionrule
\textsc{Money2} & 23.6\% & food money house bill rent stamp month today parent help pizza someone anything mom anyone \\
\sectionrule
\textsc{Job} & 31.9\% & job month rent year interview bill luck school pizza paycheck unemployment money ramen end check \\ 
\sectionrule
\textsc{Friend} & 17.0\% & friend house night mine pizza birthday thing school site place family story way movie anything \\
\sectionrule
\textsc{Student} & 23.2\% & student college final pizza loan summer university money class meal year semester story kid school \\
\sectionrule
\textsc{Time\&Family} & 23.5\% & tonight night today tomorrow someone anyone friday dinner something account family bank anything home work \\
\sectionrule
\textsc{Time} & 28.6\% & day couple anything today work pizza help pay anyone home meal food ramen someone favor \\ 
\sectionrule
\textsc{Gratitude} & 27.0\% & thanks advance guy reading anyone pizza anything story tonight help place everyone craving kind favor \\
\sectionrule
\textsc{Student} & 23.2\% & student college final pizza loan summer university money class meal year semester story kid school \\
\sectionrule
\textsc{Pizza} & 20.0\% & pizza craving hut story someone anyone domino money cheese thing request picture act title kind \\
\sectionrule
\textsc{General} & 24.1\% & time pizza year people part work hour life thing lurker story anything someone month way \\
\bottomrule
\end{tabular}
} 
\vspace{-1mm}
\caption{Topics of requests identified by non-negative matrix factorization along with their success rate (SR). Note that the average success rate is 24.6\%. Due to space limitations, we only display the 15 highest-scoring terms within each topic.}
\label{table:topic_modeling}
\vspace{-3mm}
\end{table}

\begin{table*}[ht!]
\centering
\begin{tabular}{ l p{0.30\textwidth} p{0.50\textwidth}}
\toprule
\textbf{Narrative} & \textbf{Terms} & \textbf{Example Post}\\ 
\midrule
Money & money now broke week until time last day when today tonight paid next first night after tomorrow month while account before long Friday rent buy bank still bills bills ago cash due due soon past never paycheck check spent years poor till yesterday morning dollars financial hour bill evening credit budget loan bucks deposit dollar current payed & ``\emph{Broke} \emph{until} \emph{next} \emph{paycheck}, Delaware. Really hungry and some pizza would be amazing right now. I had to pay to get my car repaired this \emph{week}, leaving me with little \emph{money} \emph{until} \emph{next} \emph{Friday} when I get \emph{paid} again. Some pizza would be really amazing. I would definitely pay it forward when I get \emph{paid} \emph{next} \emph{week}.''\\
\sectionrule
Job & work job paycheck unemployment interview fired employment hired hire & ``This is my first RAOP, low on money would really enjoy a pizza! Hey, my roommate and I are running low on cash. He lost his \emph{job} last week and I had to pay his month's rent, and I'm going to have to until he finds another \emph{job}. If someone could help us out with a pizza that would be great! Thanks!''\\
\sectionrule
Student & college student school roommate studying university finals semester class study project dorm tuition & ``\emph{Studying} for \emph{finals}, no time to go get food. Im \emph{studying} for my last batch of \emph{finals} before applying to \emph{college} in the fall (transfer \emph{student}, community \emph{college} path). very hungry but being broke and having no calc textbook I'm really pressed for time :(''\\
\sectionrule
Family & family mom wife parents mother husband dad son daughter father parent mum & ``Help out a \emph{Dad} please? [...] I'm flat out broke until tomorrow with no food in the house for dinner tonight.  My \emph{daughter} is 2 and we usually do a pizza and movie night every once in a while, and she's been asking about it.  I've got rent and car payment coming up, and bill collectors calling. I try to not let my \emph{wife} know exactly how bad we are when it gets like this, but she mentioned we didn't have anything for dinner tonight, and I can't get groceries until tomorrow.''\\
\sectionrule
Craving & friend girlfriend craving birthday boyfriend celebrate party game games movie date drunk beer celebrating invited drinks crave wasted invite & ``I went out with some \emph{friends} earlier in the week and ended up lending my \emph{friend} 20 bucks til he could get to an ATM. Long story short, we ended up pretty silly \emph{drunk} and crashed at different houses so he never got a chance to pay me back. I get paid tomorrow and I could definitely tough it out, but I'd love to down a few slices before I spend the night cleaning up my apartment.''\\
\bottomrule
\end{tabular}
\vspace{-1mm}
\caption{The main narratives of textual requests discovered through topic modeling, the terms used to detect them automatically (sorted by frequency) and example posts illustrating each narrative.}
\label{table:narratives}
\vspace{-3mm}
\end{table*}

\paragraph{Automatic Narrative Detection}
This initial exploration suggests that narratives differ substantially in how successful they are.
We define concise lexicons for each of the narratives to detect their presence or absence automatically.
It would be possible to use a fully unsupervised approach using the lexicons generated through NMF but we find that the topic boundaries are often not very clear and would make it much harder to interpret the results.
Instead, we use these topics as well as vocabulary from related LIWC categories (Linguistic Inquiry and Word Count; \citeauthor{tausczik2010psychological} \citeyear{tausczik2010psychological}) as inspiration to define concise lexicons for five different narratives identified through topic modeling (Money, Job, Student, Family, Craving). 
These lexicons along with example posts for each narrative are shown in Table \ref{table:narratives}.
To measure the usage of these narratives in requests we define simple word count features that measure how often a given request mentions words from the narrative lexicons.

\section{What Factors Are Predictive of Success?}
\label{sec:factors_result}

In this section, we first introduce our methods for measuring
each factor and then present results on which of them are predictive of success.

\subsection{Measuring the Factors}
\paragraph{Temporal Factors} 
We control for temporal or seasonal effects in the data by measuring the
specific months, weekdays, days of the month, or hour of the day as well as the
the month of the request since the beginning of the community, i.e. the ``community age''.

\paragraph{Politeness} 
We measure politeness by extracting all (19) individual features from the computational politeness model introduced by
\citeauthor{Danescu-Niculescu-Mizil+al:13b} (\citeyear{Danescu-Niculescu-Mizil+al:13b}).

\paragraph{Evidentiality} 
We study a simple measure of evidentiality in RAOP posts: the presence of an image link within the request text detected by a regular expression.
Many users provide evidence for their claims to be broke or injured
by providing a picture, e.g., a screenshot of their empty bank account
or a picture of their arm in a cast. Of the 84 images
in a random subsample of about 2000 posts, 86\% included
some kind of evidence (an empty fridge, a job termination letter,
the user themselves, etc.).

\paragraph{Reciprocity} 
The concept of ``paying kindness forward'' to someone other than your benefactor after being the recipient of a kind action is referred to as \emph{generalized reciprocity} in social psychology \cite{willer2013payitforward}.
We measure linguistic manifestations of generalized reciprocity by a simple binary feature based on regular expressions that indicates whether the text includes any phrases like ``pay it forward,'' ``pay it back'' or ``return the favor.''

\paragraph{Sentiment}
We extract sentiment annotation for each sentence of the request using the Stanford CoreNLP Package\footnote{\scriptsize\url{http://nlp.stanford.edu/software/}} and encode whether a request employs an above average (median) fraction of positive sentiment sentences through a binary feature (same for negative sentiment).
We further use count features based on lexicons of positive and negative words from LIWC (normalized by length) and a regular expression detecting emoticons to detect strong sentiment in text.

\paragraph{Length} 
We use total number of words in the request as a measure of length and hypothesize that longer requests will be more successful on average.

\paragraph{Status}
We measure status in three ways. 
First, we use the \emph{karma points}
(up-votes minus down-votes) that Reddit counts on link submissions and comments, which define a notion of status in the Reddit community.
Unfortunately, Reddit only publishes \emph{current} karma scores for all users. 
Since we are only interested in features available at prediction time,
we make sure to exclude all events that occurred after the request was first submitted
and recompute karma scores from the user's full submission history up to that point in time.
Second, we also measure whether or not a user has posted on RAOP before and thus could be considered a member of the sub-community.
Third, we extract the user account age based on the hypothesis that ``younger'' accounts might be less trusted.

\paragraph{Narrative}
To measure the usage of all five narratives in requests we use word count features that measure how often a given requests mentions words from the previously defined narrative lexicons.
We normalize these features by the total number of words in the request to remove length effects.
Here, we use median-thresholded binary variables for easier interpretation but use decile-coded variants in the following prediction task.

\subsection{Results}
We model the success probability of a request in a logistic regression framework that allows us to reason about the significance of one factor given all the other factors using success as the dependent variable and the textual, social, and temporal features as independent variables.

The logistic regression results are summarized in Table \ref{table:modelfit} and are discussed next. 
We use a standard Likelihood Ratio test to compute significance scores.

\newcommand{\signone}{\textsuperscript{}}
\newcommand{\sigone}{\textsuperscript{*}}
\newcommand{\sigtwo}{\textsuperscript{**}}
\newcommand{\sigthree}{\textsuperscript{***}}

\begin{table}[t]
\begin{center}
\setlength{\tabcolsep}{6pt}
\begin{tabular}{l @{\hspace{6pt}} r@{}l c} 
\toprule
\textbf{Coefficient} & \multicolumn{2}{l}{\textbf{Estimate}} & \textbf{SE} \\
\midrule
Intercept                                & $-2.02$ & \sigthree  & $0.14$ \\
\sectionrule
Community Age (Decile)                      & $-0.13$ & \sigthree  & $0.01$ \\
First Half of Month (Binary)                             & $0.22$ & \sigtwo   & $0.08$ \\
\sectionrule
Gratitude (Binary)                        & $0.27$ & \sigtwo   & $0.08$  \\
\sectionrule
Including Image (Binary)                      & $0.81$ & \sigthree    & $0.17$ \\
\sectionrule
Reciprocity (Binary)           & $0.32$ & \sigtwo    & $0.10$ \\
\sectionrule
Strong Positive Sentiment (Binary) & $0.14$ & \signone    & $0.08$  \\
Strong Negative Sentiment (Binary)  & $-0.07$ & \signone   & $0.08$  \\
\sectionrule
Length (in 100 Words)                        & $0.30$ & \sigthree & $0.05$  \\
\sectionrule
Karma (Decile)                               & $0.13$ & \sigthree    & $0.02$  \\
Posted in RAOP before (Binary)          & $1.34$ & \sigthree   & $0.16$  \\
\sectionrule
Narrative Craving (Binary)                       & $-0.34$ & \sigthree   & $0.09$  \\
Narrative Family (Binary)                    & $0.22$ & \sigone    & $0.09$ \\
Narrative Job (Binary)                    & $0.26$ & \sigtwo    & $0.09$ \\
Narrative Money (Binary)                       & $0.19$ & \sigtwo    & $0.08$ \\
Narrative Student (Binary)                   & $0.09$ & \signone    & $0.09$  \\
\bottomrule
\multicolumn{4}{l}{\scriptsize{\textsuperscript{***}$p<0.001$, 
  \textsuperscript{**}$p<0.01$, 
  \textsuperscript{*}$p<0.05$}}
\end{tabular}
\vspace{-1mm}
\caption{Logistic regression results for textual, social, and temporal features displaying parameter estimates and standard errors (SE) for all features. Statistical significance is calculated in a Likelihood Ratio test. All features except the ``student'' narrative and sentiment significantly improve the model fit. }
\label{table:modelfit}
\vspace{-3mm}
\end{center}
\end{table}

\paragraph{Temporal Factors}
For temporal features we find that seasonalities within a day and a year did not differentiate significantly between successful and unsuccessful requests. 
However, the success rate decreases significantly with community age. 
While the first 10\% of requests had an average success rate of 36.6\%, the last 10\% were only successful in 16.9\% of all cases (mostly, because the number of requests grew faster than the number of pizzas given out).
Further, requests in the first half of the month tend to be more successful (26.4\%) than in the second half (23.0\%); 
we encode this as a binary feature.

\paragraph{Politeness}
Out of all 19 politeness features we find that only gratitude is significantly correlated with success when controlling for temporal, social, and other textual features.
Thus, we only include gratitude in our final model.

Politeness is the expression of the speakers’ intention to mitigate face threats carried by certain face threatening acts toward another \cite{Brown1987PolitenessSomeUniversalsInLanguageUsage}. 
We speculate that in this controlled environment there is very little room for face-threats as the roles are very well defined: 
requesters do not offend any potential giver by asking for a pizza but do exactly what is expected;
potential givers can choose not to satisfy a particular request without face-threat (no direct interaction).
Without face-threats there is no need for politeness to cover the acts of the requester or to provide the potential giver with a face-saving way to deny the request.

\paragraph{Evidentiality}
Including an image greatly increases the chance of success (second largest parameter estimate) providing strong support for the hypothesis that proving additional evidence makes you more likely to succeed.
We attribute this to the fact that most pictures communicate need and urgency as well as establish an increased level of trust between requester and giver.

\paragraph{Reciprocity} 
We find that the simple linguistic indication of willingness to give back to the community significantly increases the likelihood of success.

This finding raises the question whether those users who claim to give back to the community actually live up to this claim.
To answer this question we restrict our data to only those requests that were actually successful.
We find that on average 5.9\% of successful users reciprocate (baseline rate).
Out of those that claim to ``pay it forward'' after receiving a pizza 9.9\% actually do.
While this seems like a disappointingly small fraction it is conceivable that many users have not yet been able to help someone else out and might still do so in the future. 
And indeed, this fraction is significantly larger than the baseline rate according to a binomial test ($p<0.01$). 

Does gratitude predict generalized reciprocity as suggested by \citeauthor{gray2012payingitforward} \shortcite{gray2012payingitforward}?
We find that users that express gratitude in their request return the favor 7.2\% of the time which exhibits only a slightly trend to be larger than the baseline 
(one-tailed binomial test, $p=0.115$).

We also note that high status (karma in top 20\%) is also positively correlated with reciprocity (one-tailed binomial test, $p<0.05$). 

\paragraph{Sentiment}
We find that sentiment stops being significantly correlated with success when controlling for the other variables. 
Similar results hold when using the fraction of sentences with positive/negative sentiment, thresholded versions of those features, other sentiment models and lexicons (LIWC) as well as emoticon detectors. 
This lack of relationship between sentiment and success may
be a masking effect, due to the correlation between
positive sentiment and other variables like
reciprocity (Pearson correlation coefficient $r=.08$) and word length ($r=.10$).

\paragraph{Length}
Longer requests are significantly correlated with success.
We attribute this to the fact that longer requests give the user the opportunity to provide more evidence for their situation.
Length is arguably the most simple and accessible feature associated with success.

\paragraph{Status}
We find account age to be strongly correlated with karma ($r=.75$).
This means that the ``senior'' users within the community tend to have high status (note that these are senior users that are still active as opposed to all senior user accounts).
Therefore, we only include the karma score as a decile-coded variable (indicating the decile in the overall karma distribution) in the model.
Status, in both in the Reddit community as well as the RAOP subcommunity, turns out to be strongly correlated with success.
Having posted before in ROAP also has a particularly strong positive effect on success.
People are more likely to help users that have contributed to the community in some form already
\cite{willer2009}.

\paragraph{Narrative}
All narratives significantly improve the fit except for the ``student'' narrative.
The ``job'', ``money'', and ``family'' narratives increase the predicted probability of success, while the ``craving'' narrative has strongly negative influence.
This provides strong support for our hypothesis that narratives that clearly communicate need (``job'', ``money'') are more successful than those that do not (``craving'').

\subsection{Interpretation}

\begin{figure}[tb]
	\centering
	\includegraphics[width=0.95\columnwidth]{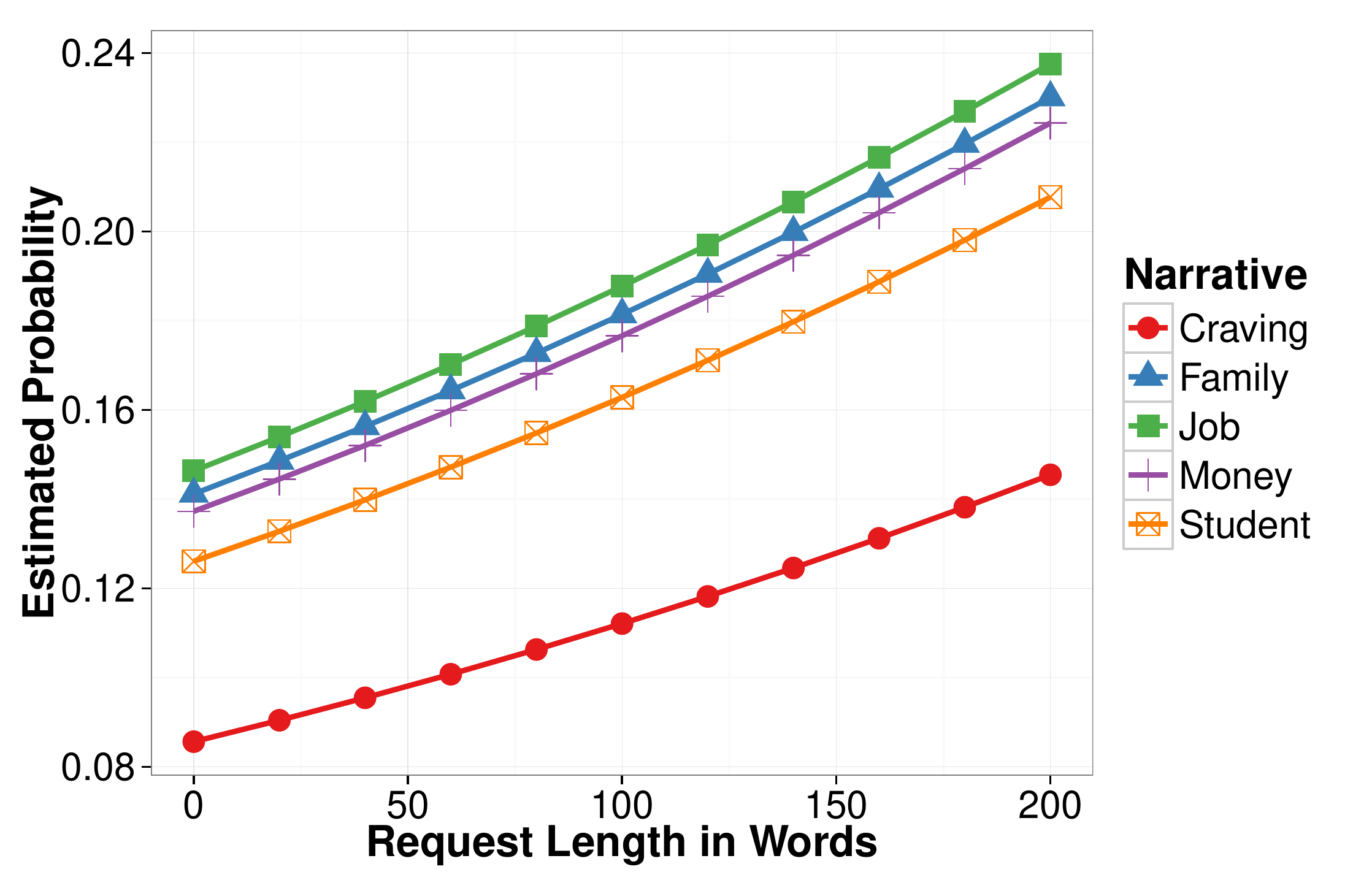}
	\vspace{-1mm}
	\caption{Estimated probability of success across request lengths for different narratives (top to bottom: Job, Family, Money, Student, Craving).}
	\vspace{-3mm}
	\label{fig:narratives_length_plot}
	\vspace{8mm}

	\centering
	\includegraphics[width=0.95\columnwidth]{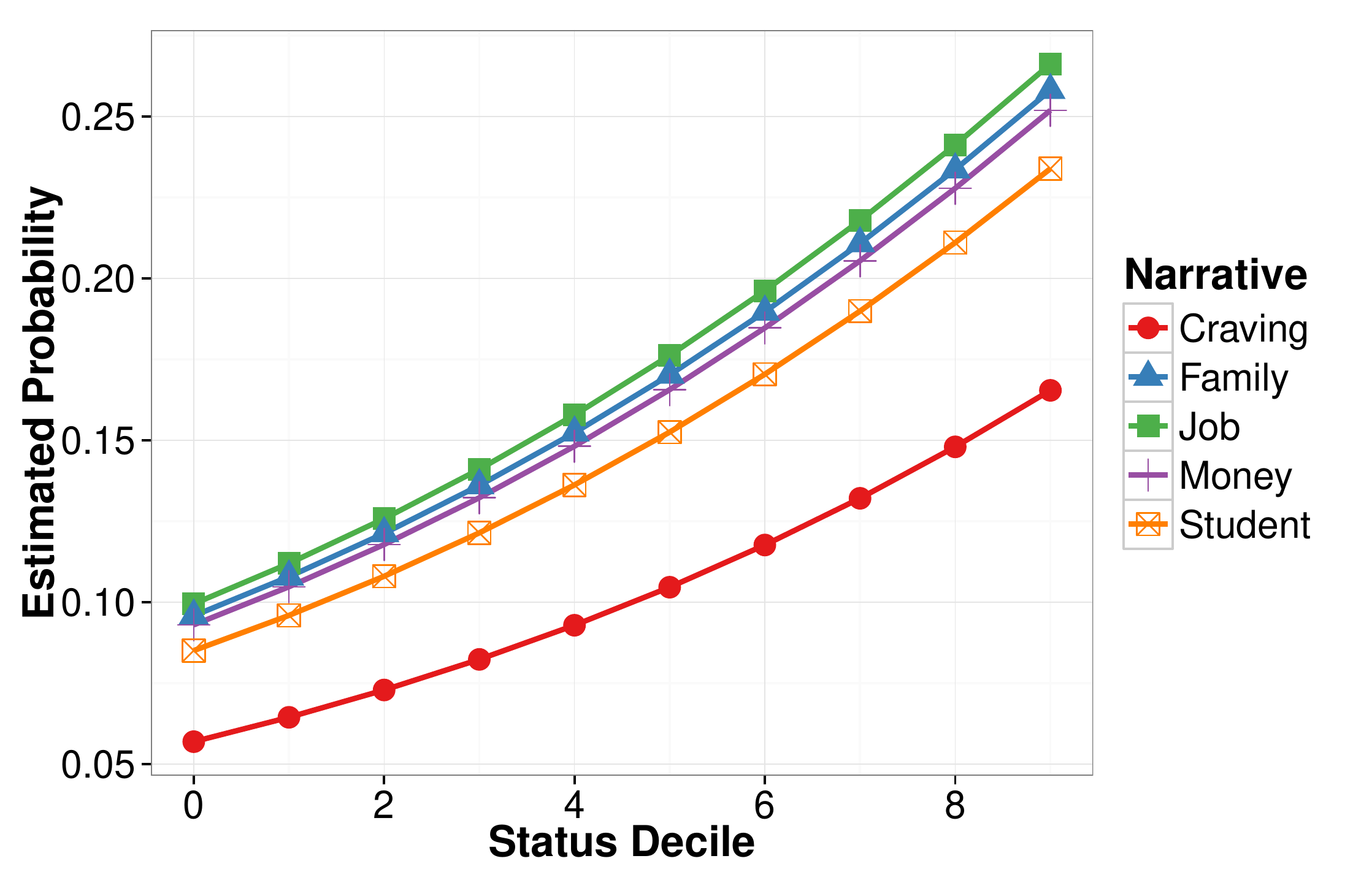}
	\vspace{-1mm}
	\caption{Estimated probability of success across status deciles for different narratives (top to bottom: Job, Family, Money, Student, Craving).}
	\vspace{-3mm}
	\label{fig:narratives_status_plot}
\end{figure}

The logistic regression parameters correspond to changes in log odds space rather than probability space and are therefore more difficult to interpret.
The change in probability space for different request lengths (median length is 74 words) is given in Figure \ref{fig:narratives_length_plot} for the different narratives (assuming all other narratives are absent).
Figure \ref{fig:narratives_status_plot} depicts the estimated success probability for different values of status (karma). 
Both plots assume that the request does not include an image, gratitude, or reciprocity claim and assumes median values for karma (or length respectively) as well as community age (thus, the success probabilities are below average).

To understand the opportunity to optimize \emph{how} one is asking for a favor and the importance to educate users about critical factors consider the following example: 
a short request (50 words) following the craving but no other narrative (assuming median karma and community age) has an estimated success probability of 9.8\%. 
Using narratives that actually display more need, say the job and money narrative instead increases the chance to success to 19.4\%, more than twice the previous probability.
Now consider another user who is smarter about how she formulates her request.
She puts in additional effort by writing more, say 150 words, and provides more evidence with a picture to support her narrative.
She also makes sure to display gratitude to the community and offers to forward a pizza to someone else once she is in a better position.
By tweaking her request in a simple way she increases her chances to 56.8\%, a dramatic increase over the former request.

\section{Is Success Predictable?}
\label{sec:prediction}

We demonstrated that textual, social and temporal factors all significantly improve the fit of a logistic regression model.
Now, we study to what degree the model is able to generalize and predict the success of unseen requests from the held-out test set. 
Because of the unbalanced dataset and the trade-off between true and false positive rate associated with prediction we choose to evaluate using the area under the receiver operating characteristic (ROC) curve (AUC) which is equal to the probability that a classifier will rank a randomly chosen positive instance higher than a randomly chosen negative one.\footnote{Note that this is closely related to the Mann--Whitney U statistic \cite{cortes2004confidence}.}
Delong's test for two correlated ROC curves is used to test for statistical significant differences in the models \cite{delong1988rocauc}.

\begin{table}[tb]
\begin{center}
\setlength{\tabcolsep}{2pt}
    \begin{tabular}{p{0.7\columnwidth} r@{}l}
    \toprule
        \textbf{Features} & \multicolumn{2}{r}{\textbf{ROC AUC}} \\
        \midrule
        Random Baseline & 0.500 & \\
        \sectionrule
        Unigram Baseline & 0.621 & \sigthree \\
        Bigram Baseline & 0.618 & \sigthree \\
        Trigram Baseline & 0.618 & \sigthree \\
        \sectionrule
        Text Features & 0.625 & \sigthree \\
        Social Features & 0.576 & \sigthree \\
        Temporal Features & 0.579 & \sigthree \\
        \sectionrule
        Temporal + Social & 0.638 & \sigthree \\
        Temporal + Social + Text & \textbf{0.669} & \textbf{\sigthree}\\
        Temporal + Social + Text + Unigram & \textbf{0.672} & \sigthree\\
        \bottomrule
    \end{tabular}
\end{center}
\vspace{-1mm}
\caption{Prediction results for logistic regression models using different sets of features. All models improve significantly upon the random baseline according to Mann--Whitney U tests ($p<0.001$).}
\vspace{-3mm}
\label{table:roc_auc}
\end{table}

Table \ref{table:roc_auc} summarizes the performance of a 
$L_1$-penalized logistic regression model \cite{friedman2010regularization} 
for different sets of features.\footnote{We also experimented with Support Vector Machines, with comparable performance.}
We include models using the standard uni-, bi- and trigram features \cite{mitra2014language} as baselines.
We further include a random baseline for comparison of ROC AUC scores.
It is important to note that there is no significant difference between our textual model with only 9 features and the uni-, bi- and trigram baselines which have orders of magnitude more features (Delong's test, $p\geq0.612$).
Both social and temporal features are predictive of success on held-out data as well.
Combining the different feature sets yields significant performance improvements from temporal (0.579) over temporal + social 
(0.638; significant improvement over individual models at $p<0.001$ in both cases), 
to all three sets of factors (0.669; significant improvement at $p<0.001$ over textual model and at $p<0.01$ over temporal + social model).
Lastly, we demonstrate that a unigram model does not significantly  improve predictive accuracy 
when combined with the proposed textual, social and temporal factors (0.672; Delong's test, $p=0.348$).
This shows that our very concise set of textual factors (narratives, evidentiality, gratitude, reciprocity, and length) accounts for almost all the variance that can be explained using simple textual models.

Although our best model (0.669) is far from perfect (1.000) all models significantly improve upon the random baseline (Mann--Whitney U test, $p<0.001$). It is worth pointing out that we are purposely dealing with a very difficult setting --- since the goal is to assist the user during request creation we do not use any factors that can only be observed later (e.g. responses, updates and comments), even though such factors have been shown to have strong predictive value \cite{etter2013launch,mitra2014language}.

\section{Does User Similarity Increase Giving?}
\label{sec:user_similarity}

  \begin{figure}[t]
    \centering
    \subfloat{
      \includegraphics[width=0.475\linewidth]{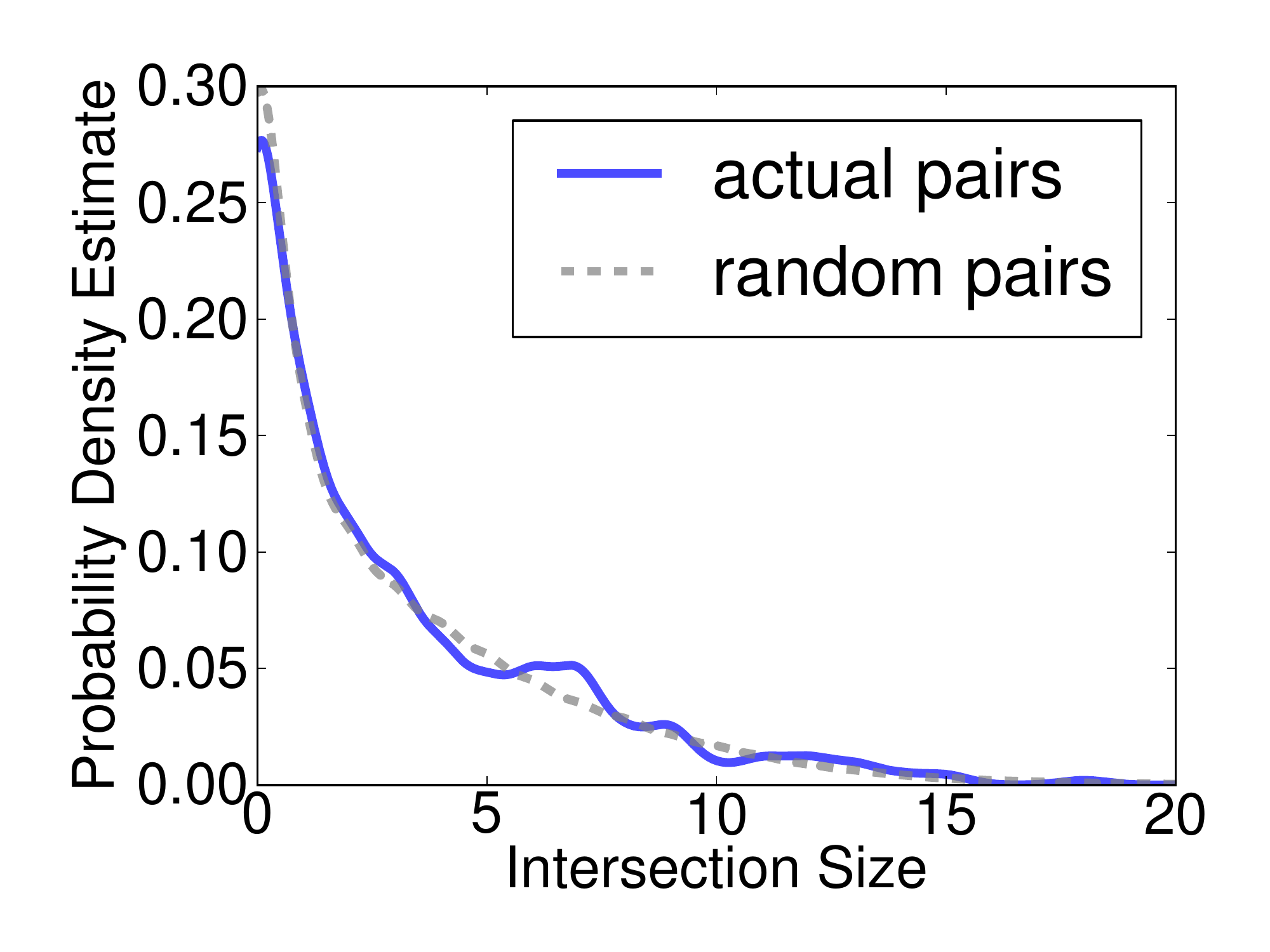}
    }
    \hfill
    \subfloat{
      \includegraphics[width=0.475\linewidth]{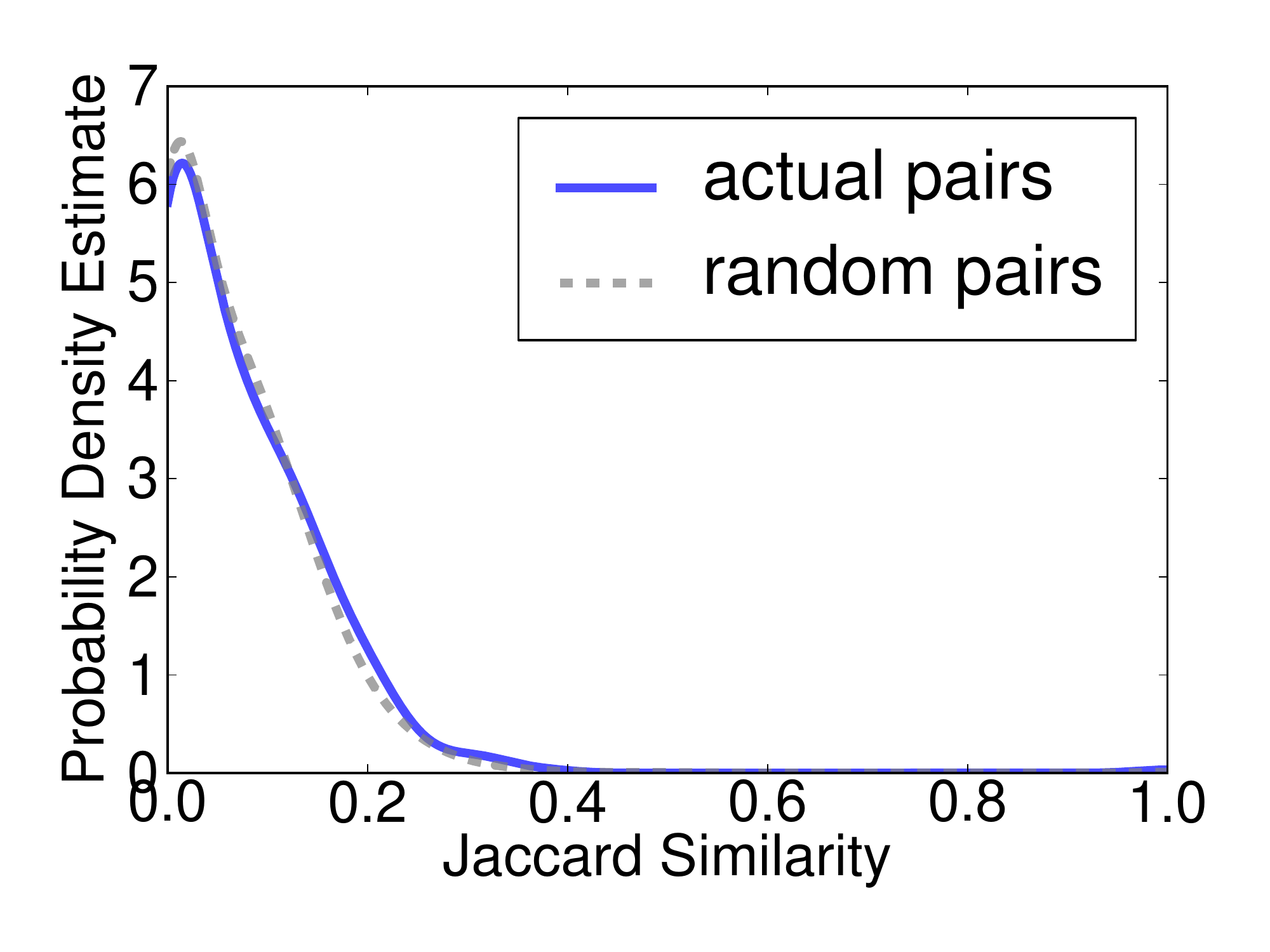}
    }
    \vspace{-1mm}
    \caption{Kernel density estimate of the similarity distribution of actual and random pairs using Intersection Size (left) and Jaccard similarity (right) as similarity metric. We do not find any significant difference between the two distributions.}
    \vspace{-3mm}
    \label{fig:user_similarity}
  \end{figure}

Social psychology literature suggests that individuals are more likely to help other individuals that are similar to themselves \cite{colaizzi1984similarity,chierco1982effects,emswiller1971similarity}.
To test this hypothesis we create a measure of user similarity by
representing users by their interests in terms of the set of Subreddits in which they have posted at least once (prior to requesting pizza),
and
employing two different similarity metrics: intersection size between the set of the giver and receiver and the Jaccard similarity (intersection over union) 
of the two.
Information about Subreddit overlap is easily accessible to users by a single click on a user's username.
We compute the similarity of giver-receiver pairs and compare it with that of random pairs of other givers and receivers (pairs that did \emph{not} occur in the data).
The latter pairing is equivalent to a random rewiring of edges in the bipartite graph between givers and receivers which we use as a null model: 
if users are indeed more likely to give to other users that are similar to them we expect the similarity of actual giver-receiver pairs in our dataset to be significantly larger than the similarity of the randomly rewired pairs.

The resulting similarity distributions for both actual and random pairs for both metrics are shown in Figure \ref{fig:user_similarity} (kernel density estimate using a Gaussian kernel with bandwidth $0.5$ for intersection size and $0.03$ for Jaccard). 
The similarity distributions for actual and random pairs match very closely, a finding that is robust across both choices of similarity metrics.
Thus, we conclude that we do not find any evidence that user similarity, at least in terms of their interest and activity as measured here, has a significant effect on giving.
They may be other indicators of similarity including geography,\footnote{We extracted location entities from the pizza requests but found them to be very sparse.} similar life situations or similar language use. 
In future work we plan to investigate other types of user similarity and their effect on helping behavior in online communities.
Note that we do not include user similarity as a feature in the logistic regression model above since we only observe givers for a small subset of requests.

\section{Conclusion}
\label{sec:conclusion}

\begin{table*}[ht]
\begin{center}
    \begin{tabular}{p{0.13\textwidth} p{0.35\textwidth} p{0.47\textwidth} }
    \toprule
        \textbf{Category} & \textbf{Prediction by Literature} & \textbf{Finding of This Case Study} \\
        \midrule
        Gratitude & A person experiencing gratitude is more likely to behave prosocially towards their benefactor and others \cite{tsang2006gratitude,bartlett2006gratitude,mccullough2001gratitude}. & \cmark \quad Studies on gratitude typically focus on the gratitude experienced by the benefactor. We find that gratitude can be paid ``forward'' before the request becomes fulfilled and that expressions of gratitude by the requester significantly increase their chance of success. However, we find no evidence that politeness, more generally, has a statistically significant impact on success in our case study.\\
        \sectionrule
        Reciprocity & People are more likely to help if they received help themselves \cite{wilke1970obligation}. The concept of paying kindness forward (rather than back) is known as ``generalized reciprocity'' in psychology \cite{willer2013payitforward,gray2012payingitforward,plickert2007networkreciprocity}. & \cmark \quad The language of reciprocity (``return the favor'') is used in a variety of ways to signal the willingness to give back to the community by helping out another member in the future (generalized reciprocity). Such claims are significantly correlated with higher chances of success. \\
        \sectionrule
        Urgency & Urgent requests are met more frequently than non-urgent requests \cite{yinon1987reciprocity,shotland1983emergency,colaizzi1984similarity,gore1997effects}. & \cmark \quad We find that narratives that clearly express need (job, money) are more likely to succeed that narratives that do not (craving). Additional support of such narratives through more evidence (images or text) further increased the chance of success. \\
        \sectionrule
        Status & People of high status receive help more often \cite{solomon1977status,goodman1993influence,willer2009}. & \cmark \quad We find that Reddit users with higher status overall (higher karma) or higher status within the subcommunity (previous posts) are significantly more likely to receive help.\\
        \sectionrule
        Mood/Sentiment & Positive mood improves the likelihood of helping and compliance \cite{forgas1998askingnicely,milberg1988moods}. & \xmark \quad When controlling for other textual, social, and temporal factors we do not find the sentiment of the text (not the sentiment of the reader) to be significantly correlated with success. \\
        \sectionrule
        Similarity & Persons are more likely to help other people when the similarity between them is high \cite{colaizzi1984similarity,chierco1982effects,emswiller1971similarity}. & \xmark \quad Measuring similarity as the number of subcommunities that both receiver and giver are active in reveals no significant difference between actual pairs of giver and receiver and a null model of user similarity. \\
        \bottomrule
    \end{tabular}
\end{center}
\vspace{-1mm}
\caption{A summary of predictions by literature on helping behavior in psychology compared to findings of this case study.}
\vspace{-3mm}
\label{table:helping_behavior_link}
\end{table*}

Online platforms have created a new mechanism for people to seek aid from other users.
Many online communities such as question \& answer sites and online philanthropy communities are created for the express
purpose of facilitating this exchange of help.
It is of critical importance to these communities that requests get addressed in an effective and efficient manner.
This presents a clear opportunity to improve these online communities overall as well as improving the chance of success of individual requests.
However, the factors that lead to requests being fulfilled are still largely unknown.
We attribute this to the fact that the study of \emph{how} one should ask for a favor is often complicated by large effects of \emph{what} the requester is actually asking for.
We have presented a case study of an online community where all requests ask for the very same contribution, a pizza, thereby naturally controlling for this effect and allowing us to disentangle what is requested from textual and social factors.

Drawing from social psychology literature we extract high-level social features from text that operationalize the relation between recipient and donor and demonstrate that these extracted relations are predictive of success.
We show that we can detect key narratives automatically that have significant impact on the success of the request.
We further demonstrate that linguistic indications of gratitude, evidentiality, and reciprocity, as well as the high status of the asker, all increase the likelihood of success, while neither politeness nor positive sentiment seem to be associated with success in our setting.

We link these findings to research in psychology on helping behavior (see Table \ref{table:helping_behavior_link}).
For example, our work extends psychological results on offline communities to show that
people behave pro-socially in online communities toward requestors who are of high status, display urgency,
and who offer to pay it forward. 
Other novel contributions of our work include the
finding that linguistic indications of gratitude lead to pro-social behavior,
and the result that higher status users are more likely to 
demonstrate generalized reciprocity. Our results thus offer new directions for
the understanding of pro-social behavior in communities in general, as well as
providing a basis for further analysis of success in social media systems.

We must recognize a number of limitations: 
a shortcoming of any case study is that findings might be specific to the scenario at hand.
While we have shown that particular linguistic and social factors differentiate between successful and unsuccessful requests we cannot claim a causal relationship between the proposed factors and success that would guarantee success.
Furthermore, the set of success factors studied in this work is likely to be incomplete as well and excludes, for instance, group behavior dynamics.
Despite these limitations, we hope that this work and the data we make available will provide a basis for further research on success factors and helping behavior in other online communities.

\paragraph{Acknowledgements}
We thank Christina Brandt and Jure Leskovec for many helpful discussions,
Niloufar Salehi and Tuan Nguyen for helping with computing user similarity and statistical analysis,
and Julian McAuley for the original idea to use the RAOP dataset.
We also thank Reddit user ``jimwll'' who generously provided pizza when we most needed it and the anonymous reviewers for their altruistic comments.
This work was supported in part by NSF IIS-1016909.

\small
\bibliographystyle{aaai}

\end{document}